\documentclass[final]{cvpr}

\usepackage{times}
\usepackage{epsfig}
\usepackage{graphicx}
\usepackage{amsmath}
\usepackage{amssymb}
\usepackage{xcolor}
\usepackage{booktabs}       
\usepackage{subcaption}
\usepackage{comment}
\usepackage{multirow}

\usepackage[pagebackref=true,breaklinks=true,colorlinks,bookmarks=false]{hyperref}

\begin{document}

\title{PatchNet - Short-range Template Matching for Efficient Video Processing}

\author{Huizi Mao\\
Stanford University\\
{\tt\small huizi@stanford.edu}
\and
Sibo Zhu\\
MIT\\
{\tt\small sibozhu@mit.edu}

\and
Song Han\\
MIT\\
{\tt\small songhan@mit.edu}

\and
William J. Dally\\
Stanford University  \\
NVIDIA Corporation.\\
{\tt\small dally@stanford.edu}
}

\maketitle

\begin{abstract}
   Object recognition is a fundamental problem in many video processing tasks, accurately locating seen objects at low computation cost paves the way for on-device video recognition.
   We propose PatchNet, an efficient convolutional neural network to match objects in adjacent video frames.
   It learns the patchwise correlation features instead of pixel features. 
   PatchNet is very compact, running at just 58MFLOPs, $5\times$ simpler than MobileNetV2.
   We demonstrate its application on two tasks, video object detection and visual object tracking. On ImageNet VID, PatchNet reduces the flops of R-FCN ResNet-101 by $5\times$ and EfficientDet-D0 by $3.4\times$ with less than 1\% mAP loss. On OTB2015, PatchNet reduces SiamFC and SiamRPN by $2.5\times$ with no accuracy loss. Experiments on Jetson Nano further demonstrate $2.8\times$ to $4.3\times$ speed-ups associated with flops reduction. Code is open sourced at \href{https://github.com/RalphMao/PatchNet}{https://github.com/RalphMao/PatchNet}.
  
\end{abstract}

\section{Introduction}
Recognizing objects in video is an important yet costly task. Many real-world applications, such as autonomous driving and video surveillance, require efficient processing of video data. Deep convolutional neural network (CNN) models are a powerful tool for recognition tasks. However, CNN models are computationally demanding, making real-time processing infeasible on embedded devices. 

As a result, many methods have been proposed to exploit the short-range correspondence that exists in many videos~\cite{zhu2017deep,mao2018catdet,luo2019detect,wang2019fast,liu2019looking}.
These methods extract explicit or implicit correspondence between frames, such as optical flow~\cite{zhu2017deep}, object tracking~\cite{mao2018catdet,luo2019detect}, motion fields in compressed video learned with LSTM~\cite{wang2019fast}. Discovering cheap and reliable correspondence in video greatly reduces processing cost while preserving prediction quality. 

\begin{figure}[h!]
    \centering
    \includegraphics[width=0.47\textwidth]{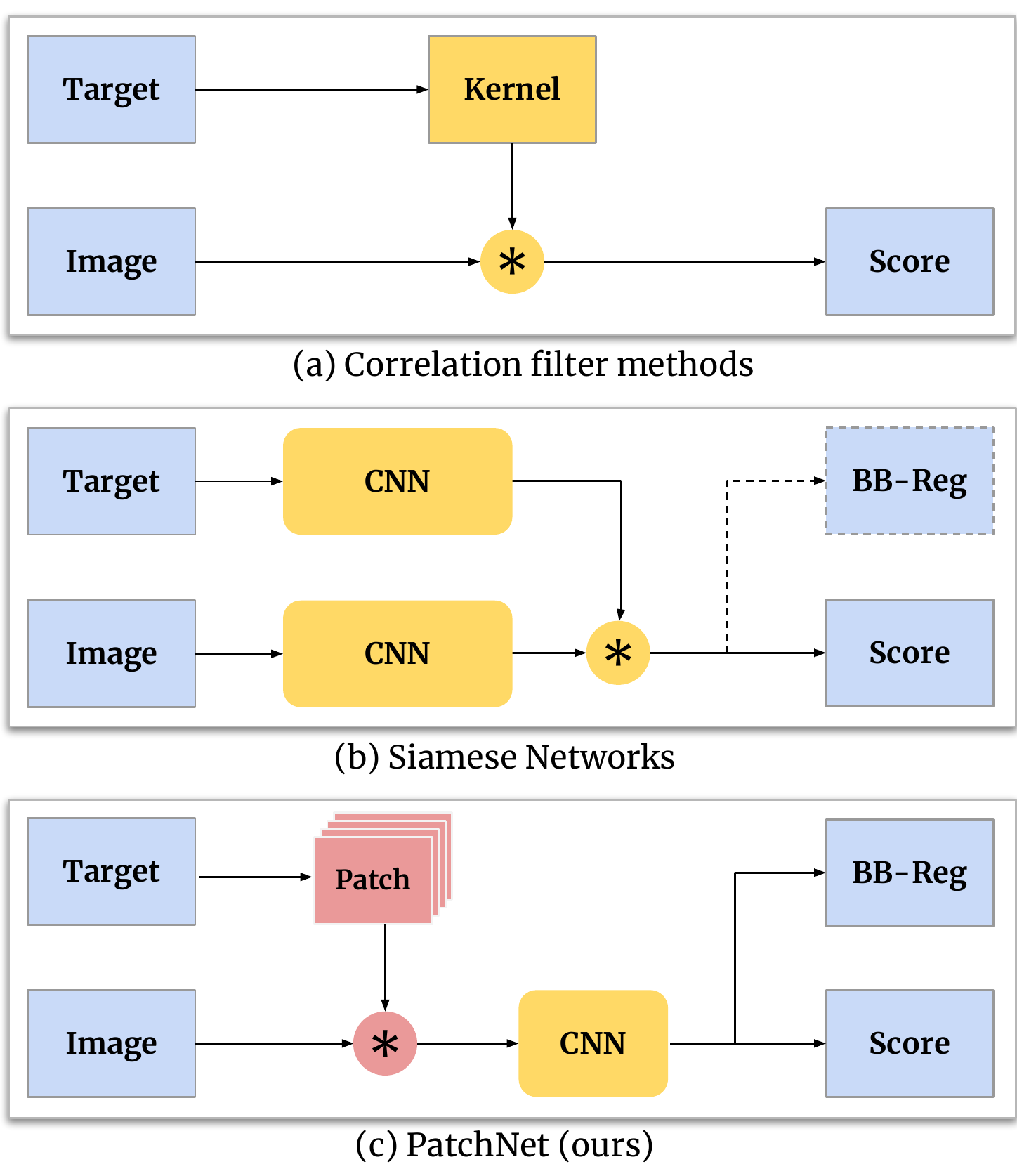}
    \caption{For short-range template matching, PatchNet combines the simplicity of correlation filter methods and the learnability of Siamese networks. PatchNet works by fitting a very efficient CNN on a patch-wise correlation map instead of image pixels. 
    }
    \label{fig:comparison}
    \vspace{-5pt}
\end{figure}

On the other hand, the recent progress on compact CNN models~\cite{howard2017mobilenets,tan2019efficientnet} has placed stricter computation budget for temporal correspondence. The budget for a CNN model has been pushed from $\sim$10GFLOPs of VGG~\cite{simonyan2014very} and ResNet~\cite{he2016deep} to $\sim$600MFLOPs in mobile regime~\cite{radosavovic2020designing}. One may find the existing techniques no longer yield significant speed-ups, as the cost to compute correspondence is no longer negligible. As an example, the recent EfficientDet-D0~\cite{tan2019efficientdet} costs only 2.5GFLOPs, while optical flow by FlowNet-Inception~\cite{zhu2017deep} alone costs about 1.8GFLOPs, 72\% of the backbone itself.


Intuitively, matching a previously seen object should be cheaper than searching for a new object. The proposed PatchNet is designed for this low-cost \textit{short-range template matching} problem. It has a low computation footprint, like correlation filter methods~\cite{henriques2014high,bolme2010visual}, but with robustness similar to deep CNN based Siamese Networks~\cite{li2018high}. As shown in Figure~\ref{fig:comparison}, instead of using a single correlation filter, PatchNet splits a template into patches and generates a multi-channel correlation map. This correlation map, treated as a feature map, is then fed into a shallow CNN model to predict object locations. In addition, we propose a hierarchical bounding box regression scheme that enables bounding box regression with just a few layers.

The contributions of this work are listed as follows:
\begin{itemize}
    \item This work describes the short-range template matching problem, and demonstrates that patch-wise correlation enables very efficient model design for this problem.
    \item For video object detection, patchNet achieves 4.9x speed-up over R-FCN ResNet-101 and 3.4x speed-up over EfficientDet-D0, with less than 1\% mAP loss.
    \item For visual object tracking, PatchNet achieves 2.6x speed-up compared to SiamRPN without loss of accuracy, or 3.9x with 1\% success score loss. 
    \item Experiments on Jetson Nano demonstrates near-linear speed-ups associated with FLOP reduction by PatchNet. 
\end{itemize}

\section{Related Work}

\textbf{Visual object tracking} (VOT) considers the problem of tracking an arbitrary object in video solely by its template in the first frame. In the past, Correlation Filter based methods such as MOSSE~\cite{bolme2010visual} and KCF~\cite{henriques2014high} have shown great success on both accuracy and performance. Recent progress of Deep Convolutional Neural Network has enabled substantial advances in this field. SiameseFC~\cite{koch2015siamese} first formulates the VOT problem in the Siamese Network paradigm. SiamRPN~\cite{li2018high} incorporates  Region Proposal Networks (RPN) from object detection, which boosts accuracy while reducing model run time. Many works follow this paradigm to further improve tracking results, such as DaSiamRPN~\cite{zhu2018distractor}, SiamMask~\cite{wang2019fast} and SiamRCNN~\cite{voigtlaender2019siam}.
Siamese Networks work well on a wide range of problems apart from VOT, such as one-shot recognition~\cite{koch2015siamese}, re-identification~\cite{chung2017two}, and similarity measuring~\cite{rao2016deep}. The foundation of all these problems is template matching, for which Siamese Network is a suitable solution.

\textbf{Video object detection} (VOD) aims to reduce the computational cost or improve accuracy of object detection with temporal information. Most VOD models are based on image object detection models, e.g., Faster R-CNN~\cite{ren2015faster}, R-FCN~\cite{dai2016r} and SSD Multi-Box~\cite{liu2016ssd}.  
Temporal information extraction, the key for efficient and accurate video recognition, can be classified as feature level or object level. Feature-level temporal information discovers correspondence across feature maps. Deep Feature Flow~\cite{zhu2017deep} employs a reduced FlowNet to estimate optical flow and extrapolates high-level CNN features. LSTM-aided SSD~\cite{liu2018mobile} adds an LSTM layer to learn implicit correspondence between multi-frame features. 
Object-level temporal information leverages previous object detections for future detections. These methods are typically associated with object tracking. Detect to Track~\cite{feichtenhofer2017detect} adds additional regression head and training loss to encourage temporally coherent detection. Detect or Track~\cite{luo2019detect} employs a SiamFC model to locate detected objects and skip detection. CaTDet~\cite{mao2018catdet} performs selected-region detection based on a multi-object tracker's predictions.

\section{Short-range Template Matching}

\subsection{Applications}
Our goal is to match a seen object in a near future frame at a very low computation cost. By matching existing objects across frames, it is possible to skip the high-cost detection computation in most frames. We list two applications that benefit from efficient short-range template matching.

\textbf{Efficient video object detection} methods exploit the redundancy in consecutive frames to reduce computation cost. For example, one can avoid detection in inter-frames via template matching~\cite{luo2019detect,yao2020deep} or feature extrapolation~\cite{zhu2017deep}. In this work, we run the base detector only on sparse keyframes and running short-range template matching on inter-frames to reduce overall workload.

\textbf{Efficient visual object tracking} can also be achieved using short-range template matching.  With a low-cost model, like PatchNet, we run expensive Siamese models on sparse key frames only.  Again, short-range template matching is run on the inter-frames.  Key frames can be selected in an online manner,   when the number of skipped frames exceeds a limit or the similarity between consecutive frames falls below a threshold.


\subsection{Template matching with patches}

Short-range template matching is very similar to visual object tracking, which is also solved as a template matching problem. 
In contrast to the visual object tracking problem where object appearance can drastically change, low-level features of an object are well preserved over a short period, as shown in Figure~\ref{fig:patch_aggregate}. Although similar features exist across multiple frames, their spatial alignment can vary due to movement, distortion, or scale change.

\begin{figure}[t!]
    \centering
    \includegraphics[width=0.43\textwidth]{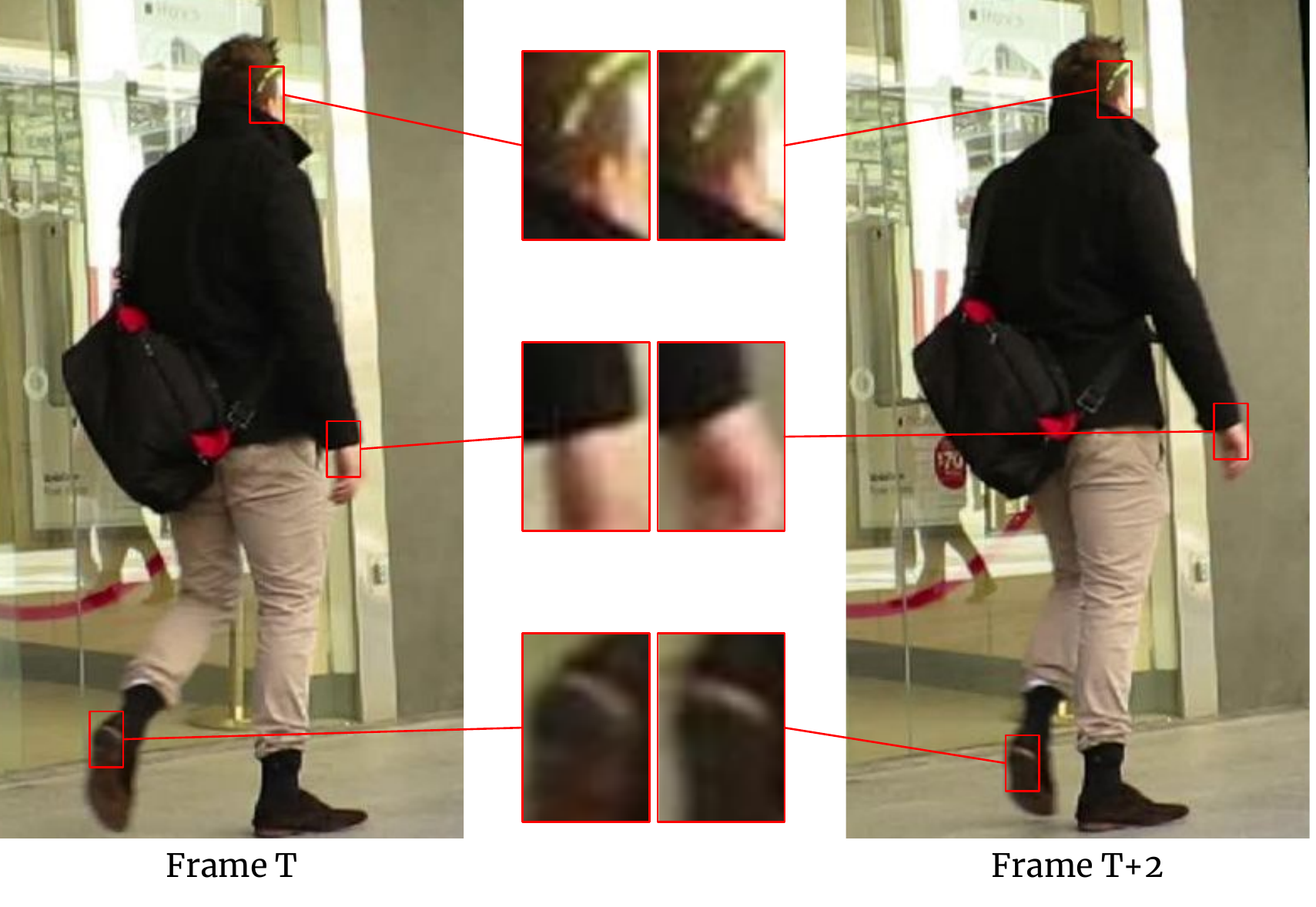}
    \caption{Similar low-level features widely exist in consecutive frames, although they may move  to different relative positions. This type of redundancy inspires a new approach to reduce computation -- learning a CNN on correlation features instead of directly on pixels.
    }
    \label{fig:patch_aggregate}
\end{figure}

Correlation filter methods~\cite{henriques2014high,bolme2010visual} have been widely applied to visual object tracking. 
A filter is created from the template and correlated with subsequent frames.
The peak correlation score locates the object.  Correlation filters are lightweight and robust to certain distortions, but especially prone to spatial variation~\cite{li2018robust,danelljan2016discriminative}. 
On the other hand, modern CNN models usually have small filters learned from data and many layers to aggregate features from low-level to high-level, and demonstrate good invariance to spatial variation.
Such robustness is attributed not only to large amount of training data and learnable parameters, but also to the great representation power of the CNN architecture. 

\begin{figure}[t!]
    \centering
    \centering
    \includegraphics[width=0.39\textwidth]{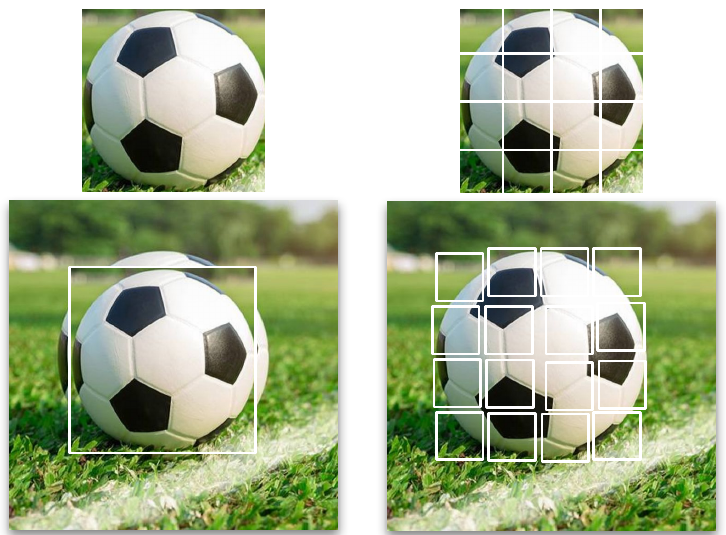}
\caption{The learned CNN model relaxes the spatial adjacency of patches and creates more robust predictions. Left: correlation filter suffers from scale change. Right: aggregated patch score from the CNN permits limited movement of patches which is more robust to scale change. }
\vspace{-10pt}
\label{fig:toycase}
    
\end{figure}
    
\begin{figure}[h!]
    \centering
    \includegraphics[width=0.45\textwidth]{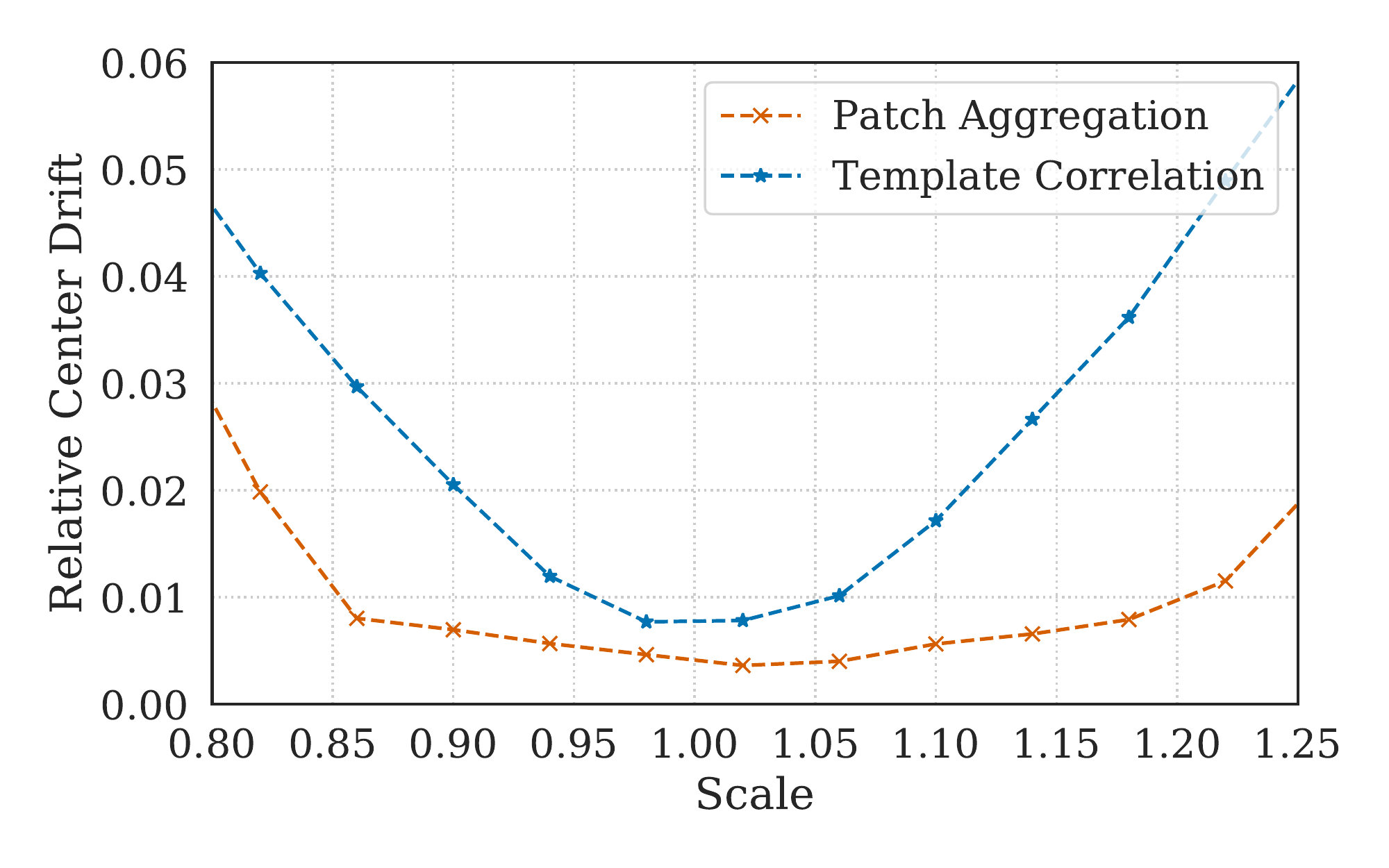}
    \caption{Localization accuracy versus scale for 1000 randomly selected ImageNet-VID objects. PatchNet works on a larger range of scales than full template correlation.}
\label{fig:patch_dist}
\end{figure}

\begin{figure*}[h!]
    \centering
    \includegraphics[width=0.85\textwidth]{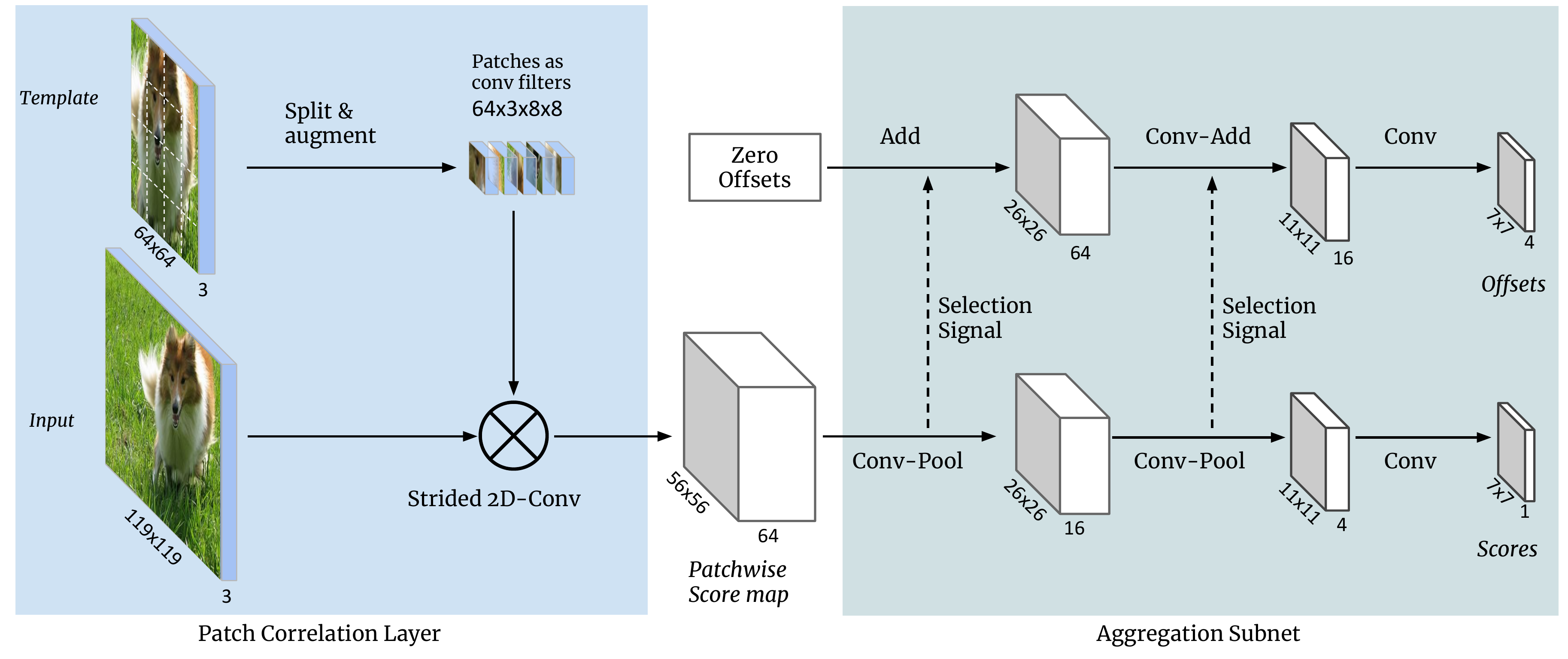}
    \caption{Overview of PatchNet. The patch correlation layer resembles a 2-D convolutional layer but with replaceable weights, which are evenly split patches from the template. It outputs a multi-channel feature map, each channel corresponding to a patch's correlation scores. The aggregation subnet, which is a 3-stage CNN, aggregates the correlation map into a single-channel feature map that localizes the object center by the peak. In addition, it has a hierarchical bounding box regression branch based on correlation information.
    }
    \label{fig:patch_aggregate}
\end{figure*}

As short-range template matching requires low-cost methods, we look for an approach with the economy of a correlation filter but with the robustness of a deep neural network.  
Inspired by the small filters and hierarchical feature extraction of a CNN, we improve upon the correlation filters by splitting a template into small patches and spatially aggregating their correlation scores. The aggregation is learned by a shallow CNN model, which combines patch-wise correlation scores into object-wise scores. This \textit{Patch Aggregation} has two main advantages. First, low level features in patches are robust to scale changes~\cite{lowe1999object}. Second, aggregation relaxes the adjacency constraint of patches using convolution and max pooling layers, creating more robust responses to spatial distortion.

The example in Figure~\ref{fig:toycase} illustrates the scale invariance of patch aggregation, where we backtrace and identify the contributing patches to the peak response score.  
We further sample 1000 objects from ImageNet VID, resize them to multiple scales and plot the center mismatch of predictions and groundtruth in Figure~\ref{fig:patch_dist}.
Full template correlation is very sensitive to scale change. Patch aggregation, on the other hand, has smaller center mismatch across a large range of scales. 
In addition to more accurate centers, we will show that patch aggregation can identify new bounding boxes at low cost, which correlation filters find difficult.

\section{PatchNet Architecture}

Figure~\ref{fig:patch_aggregate} shows the PatchNet architecture. 
PatchNet is divided into a \textit{patch correlation layer} and an \textit{aggregation subnet}. The patch correlation layer splits the template into patches and generates a 3D patch-wise correlation map. The aggregation subnet performs localization and bounding box estimation using the correlation map. Together they form a very efficient template matching model.

\subsection{Patch Correlation Layer}
In the patch correlation layer, a fixed-size template is split into a fixed number of patches. Each patch serves as a convolution filter, and together all patches form a 4D tensor as convolution weight. After that, patchwise correlation is performed by a normal 2D convolution function to generate a 3D correlation map. 
This 3D correlation map contains rich patch similarity information per channel, which is then treated as a feature map and fed into the aggregation subnet to localize the object center and regress the bounding box. 

Plain image correlation often falsely responds to background~\cite{bolme2010visual}, so does plain patch correlation.
We propose a simple yet effective learning-based method to mitigate this issue. PatchNet selects weighted Fourier-domain features rather than plain image features. It has been reported that low-frequency components are more robust to perturbations~\cite{rahaman2018spectral}. With learnable Fourier coefficients and end-to-end training, the patch correlation layer learns frequency components that are robust for recognition. In addition, Fourier transformation can be efficiently executed on both CPU and GPU. 

\begin{figure}[h!]
    \centering
  \includegraphics[width=0.45\textwidth]{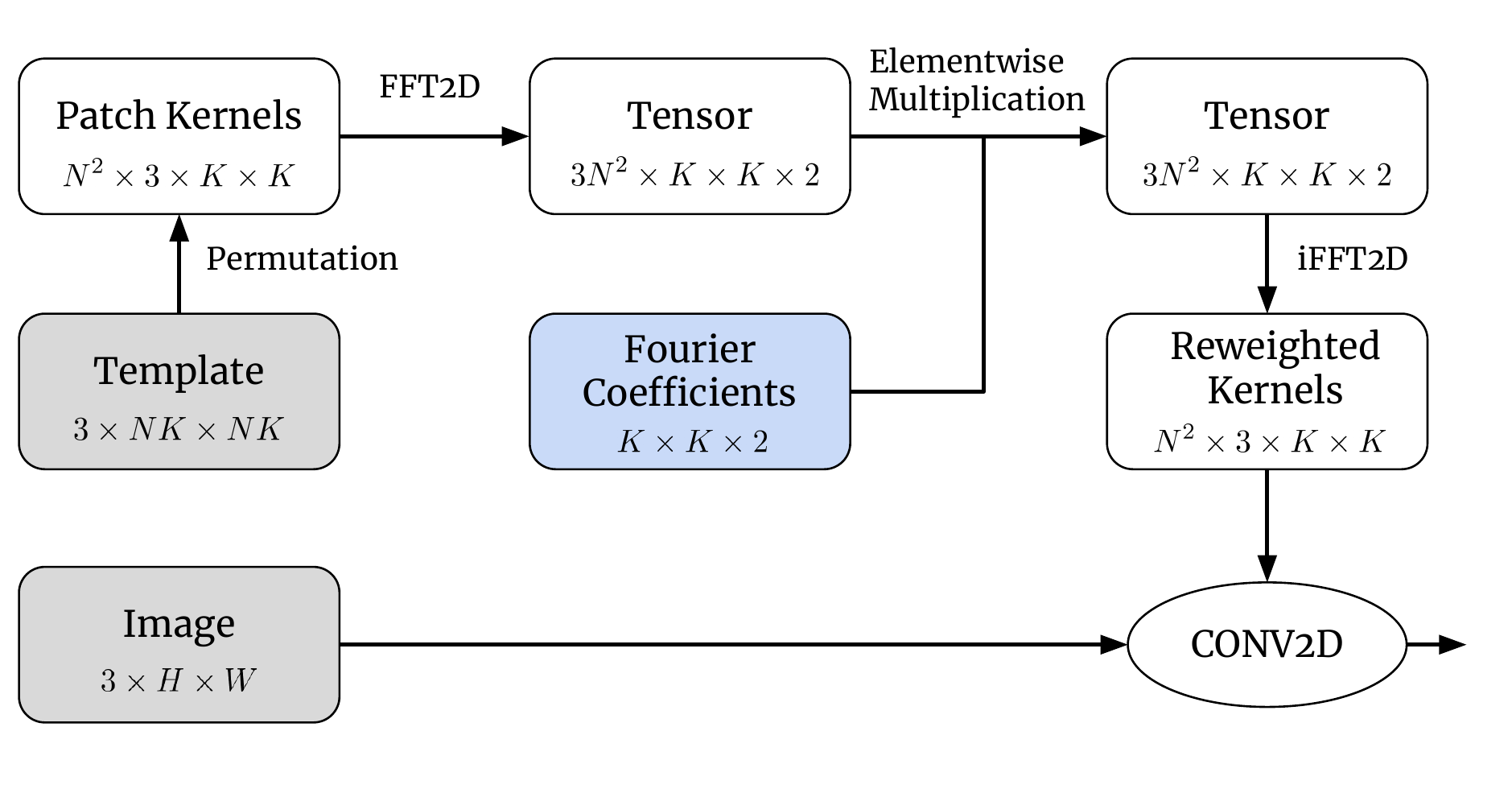}
  \vspace{-5pt}
  \caption{In the patch correlation layer, the template is split into patches,  reweighted in the Fourier domain and converted to convolutional filters.}
  \label{fig:feat_sub}
\end{figure}

Figure~\ref{fig:feat_sub} illustrates patch splitting and Fourier re-weighting. The template image, first warped to a fixed size ($KN \times KN)$, is then divided into $N^2$ patches of size $K\times K$. After they are evenly cropped out from the template, the $N^2$ patches are then transformed to the Fourier domain, multiplied by learned weights, and then converted back to the spatial domain. When $K$ has a base of 2, the Fourier transformation steps can be accelerated using an FFT with a time complexity of $O(N^2KlogK)$, which is almost negligible compared with the following convolution step's $O(HWN^2K^2)$. 

\begin{figure}[h!]

  \centering
  \begin{subfigure}{0.25\textwidth}
  \includegraphics[width=\textwidth]{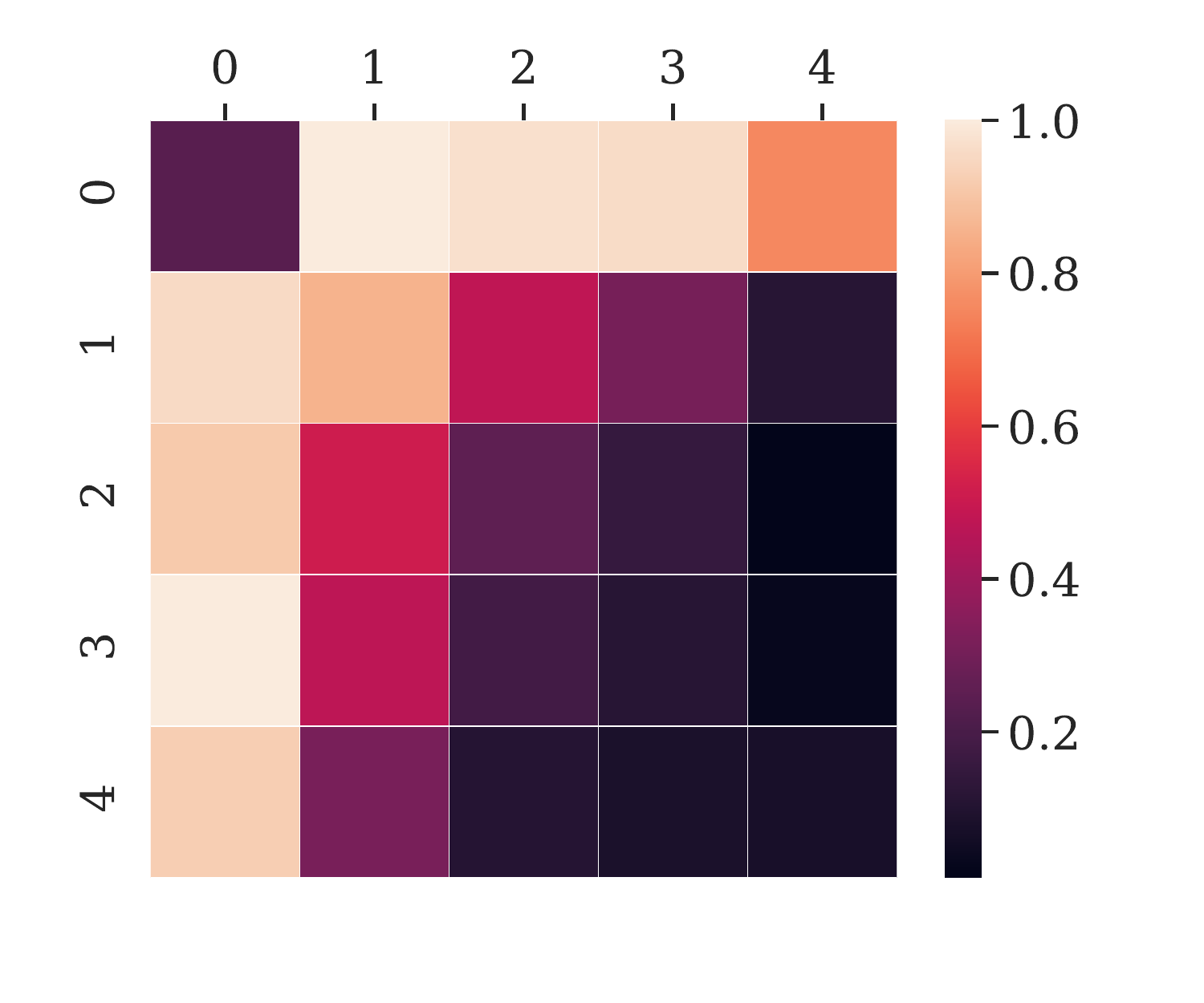}
  \end{subfigure}
  \hfill
  \begin{subfigure}{0.22\textwidth}
  \includegraphics[width=0.5\textwidth]{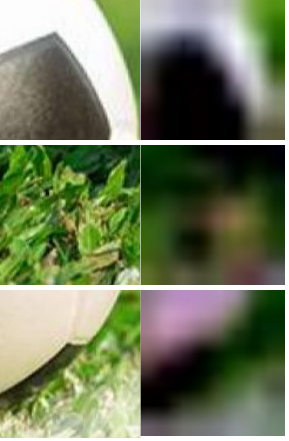}
  \end{subfigure}
  \caption{Left: learned Fourier coefficients. Right: patches before and after Fourier reweighting. The constant (DC) component and high-frequency components are both suppressed by the training process.}
  \label{fig:fcoef}
\end{figure}

The learned Fourier coefficients are visualized in Figure~\ref{fig:fcoef}. Due to the symmetry of Fourier transformation, frequency components are mirrored across axes $x=4$ and $y=4$. Therefore, we omit the mirrored weights. It is interesting that PatchNet learns to suppress both high frequencies and the constant (DC) base, implying that low-frequency information is more reliable for object recognition. 


\subsection{Aggregation Subnet}
	
The Aggregation Subnet performs target localization and hierarchical bounding-box regression. 
In our design, each convolutional layer aggregates four channels that correspond to four adjacent patches, reducing the channel dimension by $4\times$. Each pooling layer has a kernel size of 2 and stride of 2, reducing spatial dimensions by $2\times$. The output feature map has a single channel that indicates similarity to the template, similar to correlation filters.  

While there have been past works on localizing objects with just a few convolutional layers~\cite{valmadre2017end}, no existing shallow model, to our knowledge, is capable of bounding box regression. PatchNet presents a bounding box estimation strategy that works with just a few layers.
It is based on a simple observation that the patch distribution information can inform bounding box estimation, as shown in Figure~\ref{fig:toycase}. This information is hierarchically aggregated from smaller patches to larger patches, by convolution and modified max pooling.

Hierarchical bounding box regression predicts offsets of the bounding box boundary, represented by $(\Delta x_{min}, \Delta y_{min}, \Delta x_{max}, \Delta y_{max})$.
It is a three-stage procedure coupled with score aggregation. In each stage, it combines the previous stage's offset by a convolutional layer, plus the current stage's new incremental offset by the selection signal from the score pooling layer. Following this procedure, we hierarchically aggregate the per-patch offsets into box offsets.

\begin{figure}[b!]
    \centering
    \includegraphics[width=0.48\textwidth]{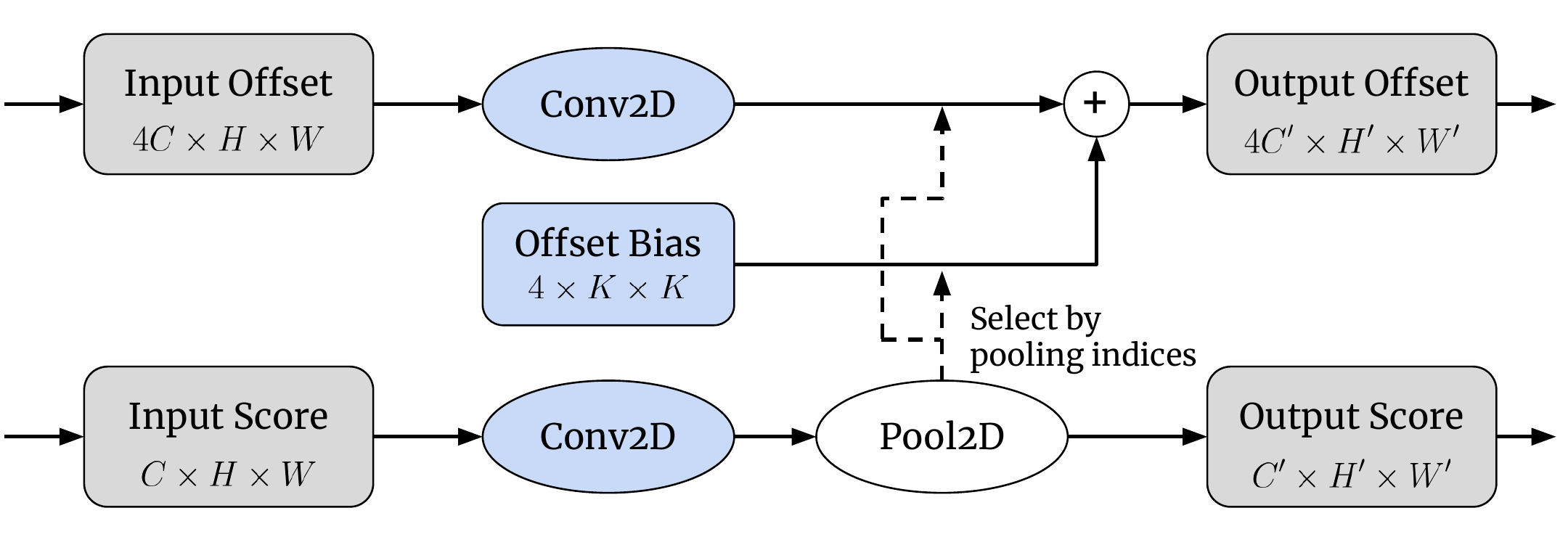}
    \caption{The building block of aggregation subnet. The lower localization path is normal conv-pool layers, while the upper bounding box regression path actively adds the partial offset from the pooling stage. 
    Input offset to the first block is set to zero. 
    }
    \label{fig:aggregate}
\end{figure}

Figure~\ref{fig:aggregate} shows the aggregation subnet's basic building block. 
The aggregation block takes two tensors as input: the score tensor represents patch-wise similarity while the offset tensor represents patch-wise boundary offsets. As each offset consists of 4 scalars, the offset tensor has four times the number of channels as the score tensor. 

Here the selection signal is the pooling indices. The intuition is that the pooling choice represents a patch shift, which contributes a learned offset to the bounding box. 
This operation first selects the offset according to the pooled score tensor, and then adds the offset bias based on its relative location inside the pooling kernel. In practice, the selection is ``soft'', i.e., a weighted combination by the magnitude of scores, instead of ``hard'' selection by indices.  Equation~\ref{eq:poolselect} describes the output offset of a pooling kernel $\mathcal{K}$, where $S$, $F$, $b$ are scores, offsets and bias, respectively.

\begin{align}
\begin{split}
    \mathbf{f}_{out} = & \sum_{\mathbf{x} \in \mathcal{K}} \hat{S} [\mathbf{x}](F[\mathbf{x}] + b[\mathbf{x}]), \\
    \quad & \text{where} \quad \hat{S} = softmax(S)
\end{split}
\label{eq:poolselect}
\end{align}

Learned aggregation is performed by convolutional layers. In the two parallel Conv2D layers, offsets and scores of neighboring patches are aggregated into larger patches, along with reduction in the channel dimension. The convolutional layer for scores is initialized with weights from the patch aggregation model. The layer for offsets is initialized so that the offset of a patch is averaged from four adjacent smaller patches.

The input offset to the first aggregation block is always set to zero. 
The final output is a single-channel response map and a 4-channel offset map. In the inference stage, we select the peak in the response map and the corresponding offset to estimate the new bounding box.


\subsection{Training}

\textbf{Objective functions}. 
We train PatchNet with a sum of localization loss and a bounding box regression loss. We adopt a simple L1 localization loss that penalizes large response scores at non-center locations. In SiamRPN, the localization task is trained as a binary classification problem with cross entropy. However, in our experiments, cross entropy loss leads to inferior localization accuracy for the shallow CNN model.  Equation~\ref{eq:centerness} describes this localization loss, where the response map is denoted by $S$ and groundtruth center location by $\textbf{x}_{gt}$. $\mathcal{D}$ is the Manhattan distance. This loss function relaxes the penalty for locations close to the center, as neighboring regions typically respond stronger than backgrounds.

\begin{align}
\label{eq:centerness}
    \mathcal{L}_{c}(\mathbf{S}, \mathbf{x}_{gt})= \sum_\mathbf{x} max (S[\mathbf{x}] - S[\mathbf{x}_{gt}] + \alpha \mathcal{D}(\mathbf{x}, \mathbf{x}_{gt}), 0)
\end{align}

For bounding box regression, the widely used smooth L1 loss is adopted, although it should be noted that output and groundtruth are both boundary offsets, different from typical bounding box regression.

\textbf{Structural sparsity constraints}. 
$N\times N$ patches are stored along the same channel dimension for efficient implementation. However, in this approach we lose the patch adjacency information. Ideally, a 4D score tensor with 4D convolution would preserve the patch adjacency, but that is beyond the capability of most deep learning libraries.
An alternative is to enforce structural sparsity in the convolution filters. We set zeros for most of the channel-wise connections except those from adjacent patches. Our experiment shows this approach reduces the gap between the training and validation losses and improves tracking robustness.

\section{Experiments}
PatchNet solves the short-range template matching problem, which is exploited to speed up video processing. 
In this section, we demonstrate the experimental results of PatchNet on video object detection (VID) and visual object tracking (VOT).
The PatchNet model is trained on the GOT-10k dataset~\cite{huang2019got}, following the training process in the SiamFC repo\footnote{ \href{https://github.com/huanglianghua/siamfc-pytorch}{https://github.com/huanglianghua/siamfc-pytorch}}. We use the same PatchNet weights for both VID and VOT benchmarks without retraining, showing that it is easy to generalize to different tasks and integrate with other models.

\subsection{Video object detection}
We select two object detection models, R-FCN ResNet-101 and EfficientDet-D0 as our baselines. R-FCN ResNet-101 is a high-precision detection model that is the baseline in many VID works~\cite{feichtenhofer2017detect,zhu2017deep}. For the ImageNet VID dataset, it consumes an average of 169GFLOPs per frame, which is computationally prohibitive on many edge devices. In contrast, EfficientDet-D0 is a state-of-the-art cost-effective detection model, consuming only 2.5GFLOPs per frame. The average number of objects per frame is 7 for R-FCN and 4 for EfficientDet on ImageNet VID.

We adopt a computation-reduction approach similar to Detect or Track~\cite{luo2019detect}. In this approach, the base detector runs on key frames and PatchNet tracks detected objects on non-key frames. The objects from key frames, together with the detection confidence scores, are preserved for a fixed number of frames unless they exceed the frame boundary.

\begin{figure}[h!]

    \includegraphics[width=0.46\textwidth]{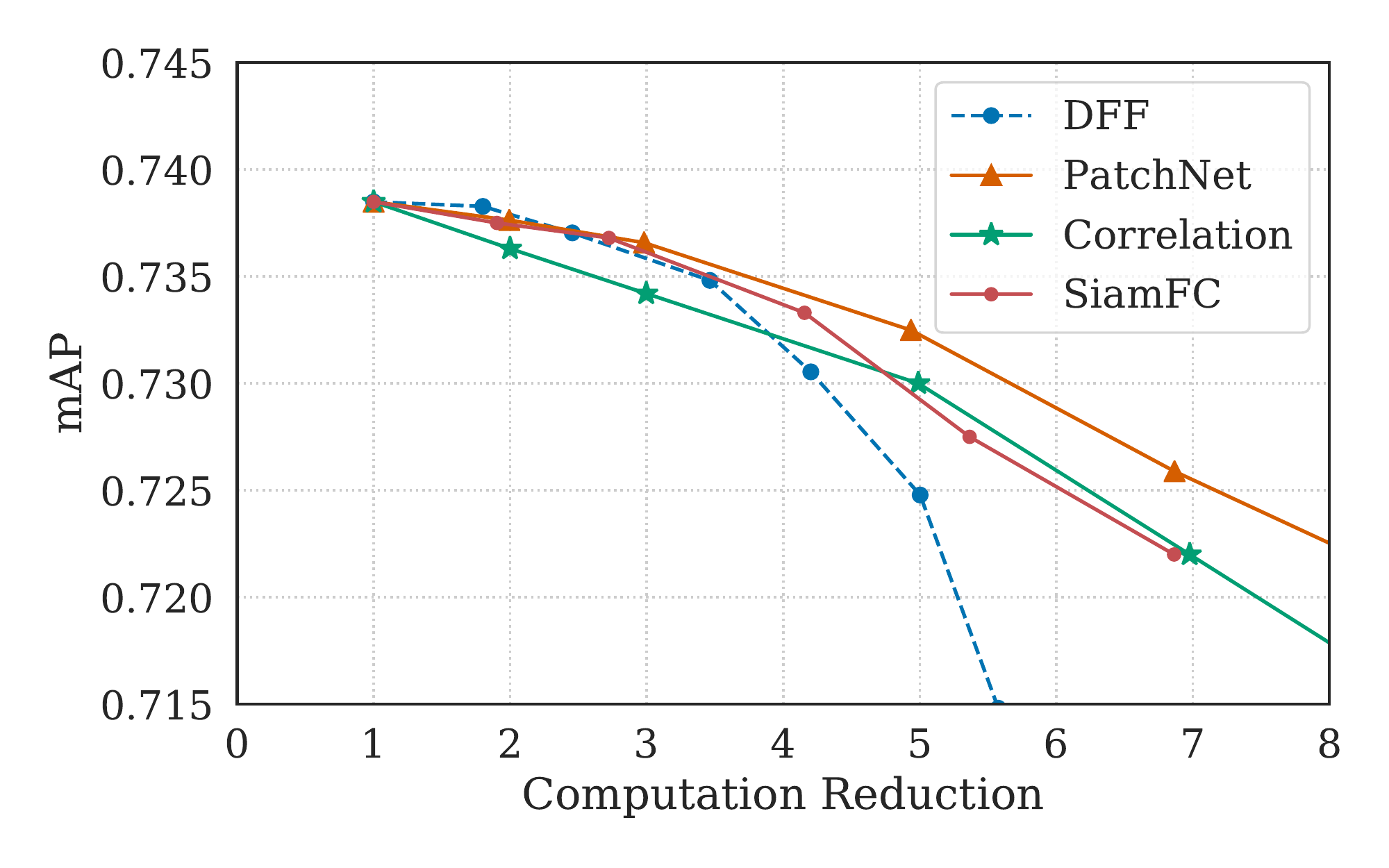}
    \vspace{-10pt}
    \caption{Complexity/accuracy trade-off with R-FCN ResNet-101 as backbone on VID dataset. Unlike DFF, PatchNet does not require retraining on video dataset, while still achieving significant speed-ups.}
    \label{fig:dffvspatch}

\end{figure}

Our results are compared against Deep Feature Flow (DFF) and other template matching methods, including full template correlation and SiamFC. DFF tries to alleviate expensive deep CNN feature extraction by propagating CNN features according to optical flow. DFF has extra overhead mostly from FlowNet, while PatchNet is much more compact. In addition, DFF requires joint training, while PatchNet does not need to be coupled with the detection model.
As shown in Figure~\ref{fig:dffvspatch}, our method generally achieves close or better speed-accuracy trade-off than other methods, especially at high amounts of computation reduction. Key results are reported in Table~\ref{tab:vid}.

\begin{figure}[h!]

    \includegraphics[width=0.46\textwidth]{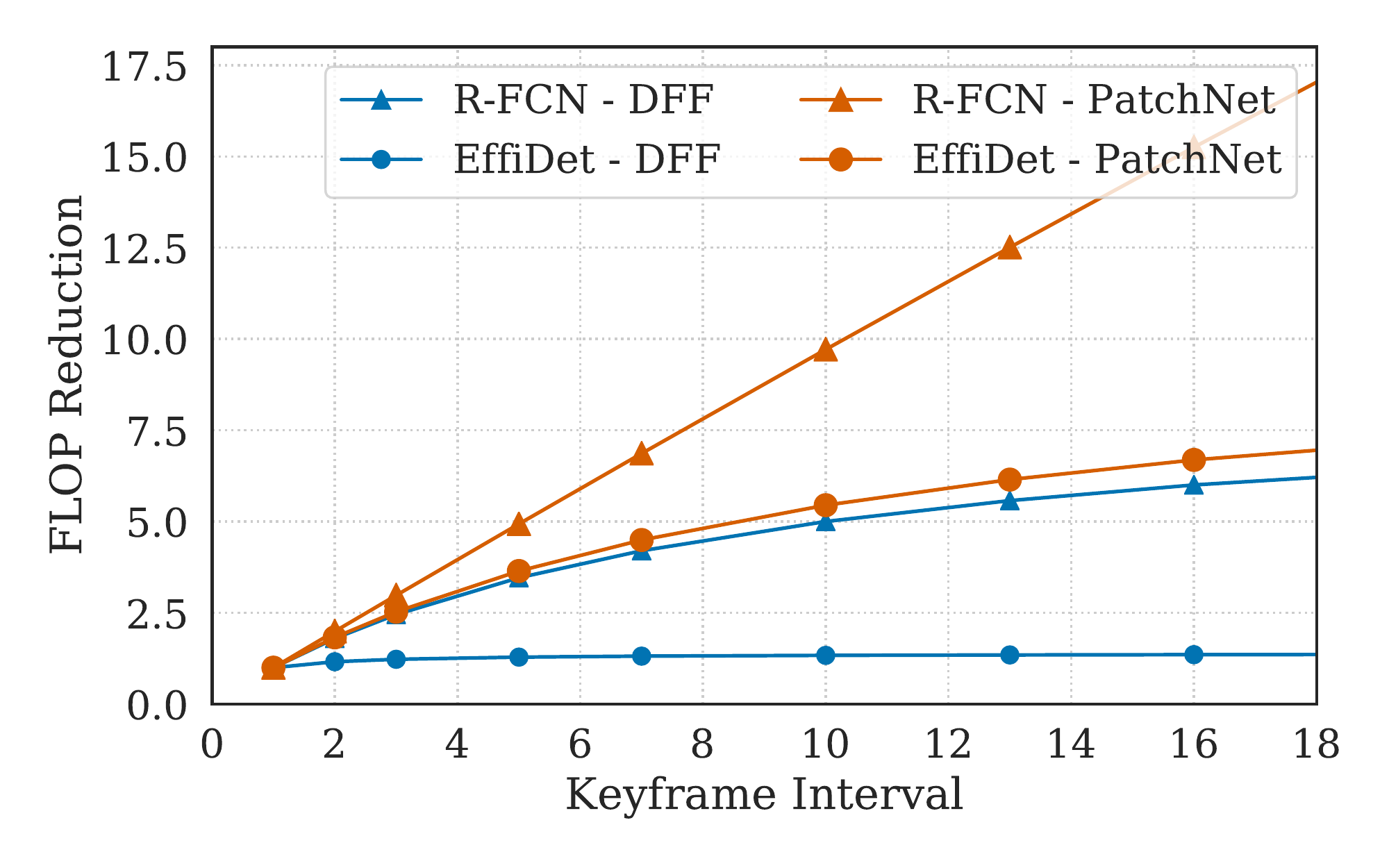}
    \vspace{-10pt}
    \caption{Projected relationship of key frame interval and model complexity reduction. PatchNet enables significant FLOP reduction for both models.}
    \label{fig:interval_flops}

\end{figure}

\begin{table}[h!]
    \centering
    \begin{tabular}{cccccc}
        \toprule
         Model & Method & mAP & FLOPs & Reduction\\
         \midrule
          \multirow{2}{*}{R-FCN }& Baseline & 0.738 & 169G & \textbackslash \\
          \multirow{2}{*}{ResNet-101} & DFF & 0.725 & 34.9G & 4.8$\times$ \\
           & PatchNet & 0.731 & 34.2G & \textbf{4.9$\times$} \\
         \midrule
         EfficientDet & Baseline & 0.597  & 2.5G & \textbackslash \\
         -D0 & PatchNet & 0.589 & 0.73G & \textbf{3.4$\times$} \\
        \bottomrule
    \end{tabular}
    \caption{PatchNet achieves better FLOPs/mAP trade-off than DFF. DFF's results are obtained with model checkpoints provided in the official GitHub repo.}
    \label{tab:vid}
    \vspace{-5pt}
\end{table}

\begin{table*}[h!]
    \centering
    \begin{tabular}{cccccc}
    \toprule
         Baseline model & Keyframe interval & Success score & Precision score  & FLOPs & Reduction\\
         \midrule
       & 1 & 0.575 & 0.766 &  2.7G & \textbackslash \\
      SiamFC  & 2.7 & 0.573 & 0.784  & 1.1G & 2.5$\times$\\
        & 3.6 & 0.562 & 0.776  & 0.8G & 3.2$\times$\\
      \midrule
       & 1 & 0.630 & 0.837  & 4.1G & \textbackslash\\
      SiamRPN-2x & 2.7 & 0.629 & 0.839 & 1.6G & 2.6$\times$ \\
        & 4.2 & 0.620 & 0.841 & 1.1G & 3.9$\times$\\
    \bottomrule
    \end{tabular}
    \vspace{5pt}
    \caption{Results on OTB2015 dataset, with SiamRPN and SiamFC as baseline models. The SiamRPN-2x model has 2x channels compared to the original paper. When keyframe interval is 1, we run the default baseline model.}
    \label{tab:vot}
    \vspace{-5pt}
\end{table*}

We implemented EfficientDet for ImageNet VID and show that PatchNet can still yield meaningful speed-ups.
DFF for EfficientDet is not reported in Table~\ref{tab:vid}, because even the smallest FlowNet-Inception is about the same complexity as EfficientDet-D0. On contrary, PatchNet can speedup the very compact EfficientDet-D0 almost linearly, as indicated in Figure~\ref{fig:interval_flops}.

\subsection{Visual object tracking}

We select SiamFC and SiamRPN as our baseline models. SiamFC is the predecessor of recent advances of Siamese models for VOT. It has an AlexNet backbone for feature extraction and computes similarity by cross-correlation. Due to lack of bounding box regression, multi-scale evaluation is necessary to handle scale change. SiamRPN adds a RPN head for bounding box regression, eliminating multi-scale evaluation while improving tracking robustness. The SiamRPN model \footnote{\href{https://github.com/huanglianghua/siamrpn-pytorch}{https://github.com/huanglianghua/siamrpn-pytorch}} uses double-channel AlexNet as the backbone, therefore we denote it by SiamRPN-2x throughout this section.

We test PatchNet with the skip-frame scheme. Key frames are selected in an online manner: when PatchNet predicts a confidence score lower than a threshold or when the number of inter-frames exceeds a limit. 
Our results are reported in Table~\ref{tab:vot}. The skip-frame method reduces average workload of both SiamFC and SiamRPN by about 3x. For both models, the success score drops by about 1 percent while the precision score is slightly improved. 



We also evaluate our methods on UAV123 ~\cite{mueller2016benchmark}, a high-resolution aerial-view dataset that is larger and more realistic than the previous OTB2015 dataset. As shown in Table~\ref{tab:uav_exp}, our skip-frame method maintains the success score and slightly improves the precision score when we run the base model every 3 frames. With a tolerance of 1 percent accuracy drop, we can further increase the keyframe interval to 5.  The results show that PatchNet can generalize to different datasets without additional training.

Table~\ref{tab:vot_comparison} shows PatchNet's advantage as an inter-frame tracker compared with other methods. All methods have key-frame intervals of roughly 4 frames. The dummy tracker does not predict any new location in the inter frames, therefore suffers from significant accuracy drop. Kernelized Correlation Filter (KCF)~\cite{henriques2014high} is a very fast correlation filter method, but it has inferior accuracy compared with PatchNet. In our experiment we use OpenCV's KCF implementation. Patch Aggregation is PatchNet without Fourier reweighting and bounding box regression.  It achieves performance intermediate between KCF and the full PatchNet.

\begin{table}[h!]
\begin{tabular}{cccc}
\toprule
Keyframe & \quad Success \quad \quad & Precision \quad & Reduction\\
interval & score  &  score & \\
\midrule
1  & 0.599 & 0.770 & \textbackslash \\
3 & 0.600 & 0.780 & 2.8$\times$ \\
5 & 0.589 & 0.777 & 3.6$\times$ \\
\bottomrule
\end{tabular}
\caption{Results on UAV dataset with SiamRPN-2x as the baseline model.}
\label{tab:uav_exp}
\end{table}

\begin{table}[h!]
\begin{tabular}{cccc}
\toprule
Skip-frame & Keyframe & Success & Precision  \\
 method & interval & score & score  \\
\midrule
Original & 1 & 0.630 & 0.837 \\
Dummy & 4.0 & 0.204 & 0.152 \\
KCF & 3.7 & 0.584 & 0.770 \\
Patch Aggregation & 3.7 & 0.609 & 0.815 \\
PatchNet & \textbf{4.2} & \textbf{0.620} & \textbf{0.841} \\
\bottomrule
\end{tabular}
\caption{Comparison with other choices for the skip-frame method. Results obtained on OTB2015, with SiamRPN-2x as the keyframe model.}
\label{tab:vot_comparison}
\end{table}

\subsection{Ablation study}
We analyze the design choices that affect PatchNet's ability to capture short-range correspondence, as measured by the miss rate of groundtruth objects that are 10 frames apart on the ImageNet VID dataset. An IoU score below 0.7 is decided as a miss, which is stricter than the object detection problem. We name this experimental setting as short-range ImageNet VID benchmark.

\textbf{Patch size} affects the representation and generalization power of PatchNet. A patch size of 1 is unlikely to work, as pixel correlation does not yield meaningful information; on the other hand, large patch sizes will incur extra computation overhead, which is unwanted for edge applications. A patch size as large as the object will convert PatchNet to a correlation filter method.

\begin{figure}
    \centering
    \includegraphics[width=0.48\textwidth]{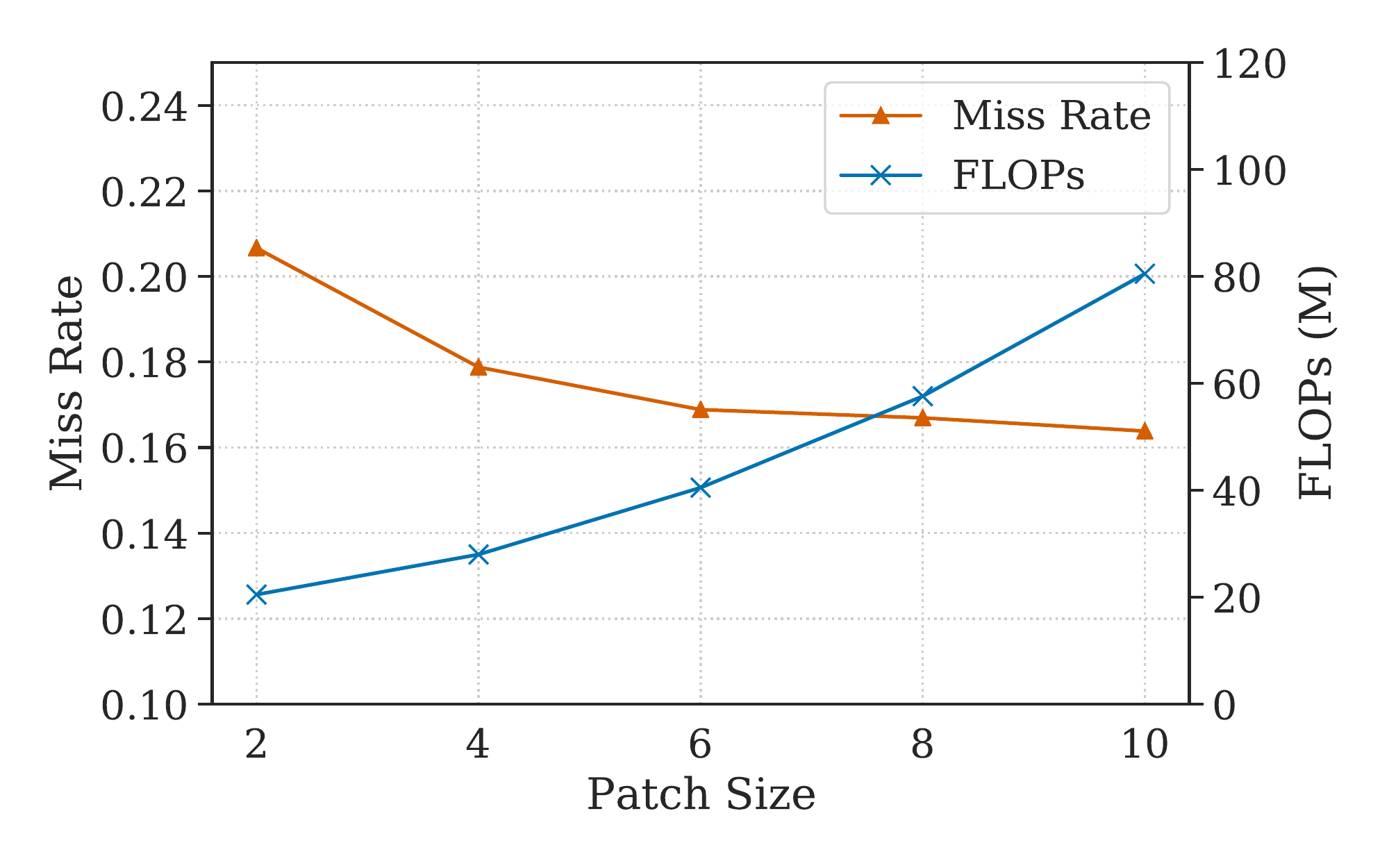}
    \vspace{-10pt}
    \caption{Accuracy and complexity of different patch sizes on the short-range ImageNet VID benchmark.}
    \label{fig:patchsize}

\end{figure}

Figure~\ref{fig:patchsize} shows the impact of patch size on accuracy and complexity. The number of patches is fixed to keep the aggregation subnet unchanged. Larger patch sizes yields diminishing return on accuracy, while greatly increasing computation cost. 

\textbf{Fourier reweighting} and \textbf{hierarchical bounding box regression} are two novel components in PatchNet. Table~\ref{tab:ablation_component} shows their impact on accuracy and overall computation cost. 
Starting from Patch Aggregation, Fourier basis weighting gives a 1.7\%  miss rate improvement while increasing computation cost by only 1 MFLOPs. Such a small overhead is attributed to Fast Fourier Transformation for base-2 patches.
Hierarchical bounding box regression, on the other hand, gives a 3.2\% miss rate improvement while increasing computation footprint by 12.4 MFLOPs.

\begin{table}[t!]
\centering
\begin{tabular}{ccc}
\toprule
Method & Miss rate & FLOPs\\
\midrule
Patch Aggregation & 0.208 & 44.2M \\
w/ Fourier reweighting & 0.199 & 45.2M \\
w/ BB regression & 0.184 & 56.6M \\
PatchNet & 0.167 & 57.6M \\
\bottomrule
\end{tabular}
\caption{Analysis of Fourier reweighting and hierarchical bounding box regression on the short-range ImageNet VID benchmark.}
\label{tab:ablation_component}

\end{table}

\subsection{Speed on edge device}

We evaluate PatchNet on an NVIDIA Jetson Nano to measure its performance on a typical edge device.
Efficient model execution on mobile platforms has become a focus in many recent works~\cite{howard2017mobilenets,tan2019efficientnet}. In this section, we demonstrate that PatchNet is able to accelerate all aforementioned models on a mobile platform.

\begin{figure}[t!]
    \centering
    \includegraphics[width=0.47\textwidth]{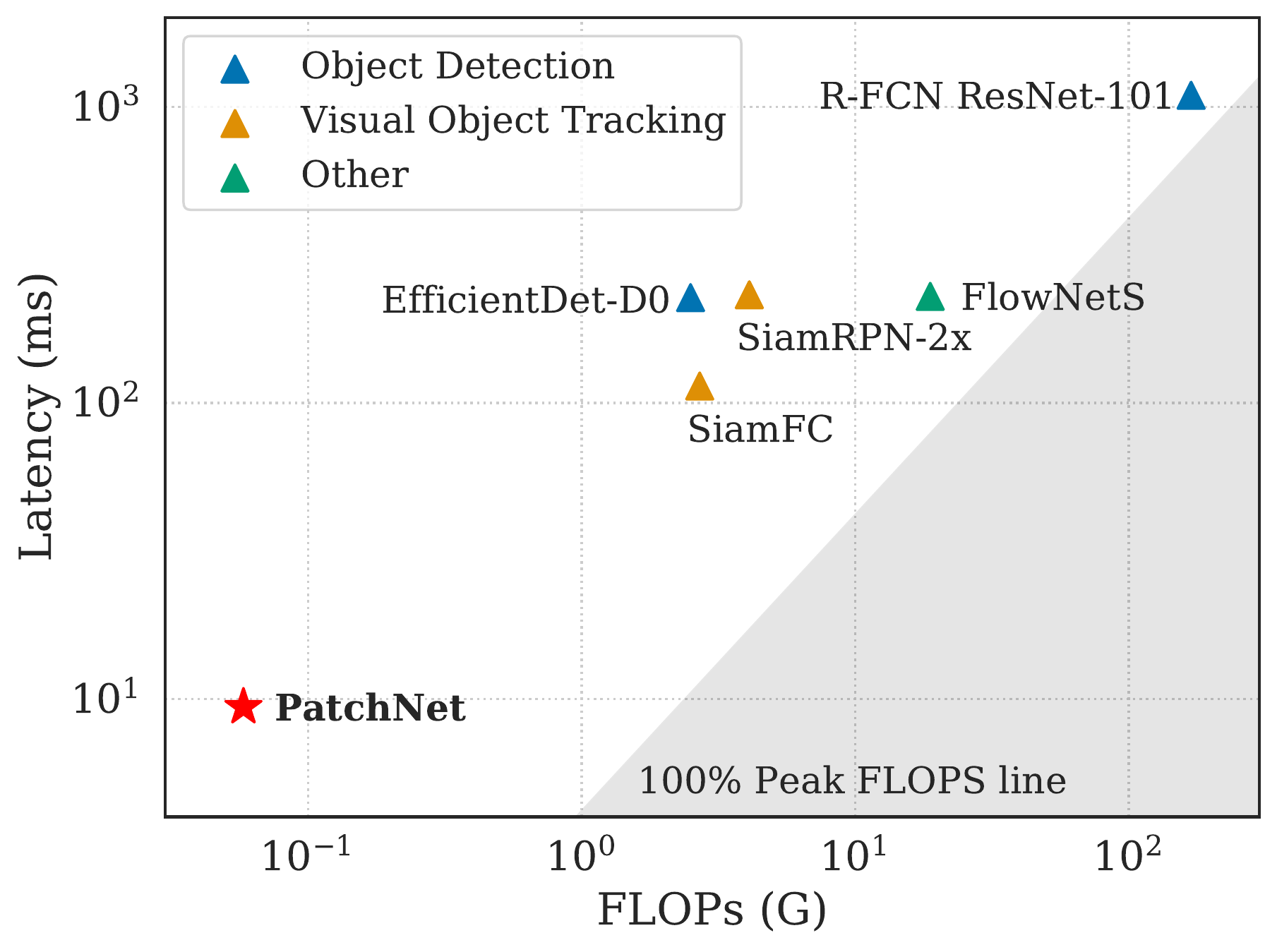}
    \vspace{-5pt}
    \caption{Single-frame inference speed on Jetson Nano. PatchNet is orders of magnitude faster than other models, making it an ideal inter-frame model. }
    \label{fig:speed_flops}
\end{figure}

All models listed in Figure~\ref{fig:speed_flops} are tested with PyTorch-1.4 on Jetson Nano with a quad-core ARM Cortex-A57 CPU and an 128-core Maxwell GPU, with a batch size of 1. 
As shown in the figure, PatchNet has orders of magnitude lower on-device latency and consumes orders of magnitude fewer FLOPs than the other models. 

In Table~\ref{tab:nano}, we list the speed-ups in accordance to Table~\ref{tab:vid} \& \ref{tab:vot} in previous sections. PatchNet achieves 2.8 - 4.3$\times$ speedup on four listed models (excluding data IO time). Model optimization with TensorRT may further improve the speed, as already shown on many edge device applications~\cite{hadidi2019characterizing,jo2020benchmarking}.

\begin{table}[t!]
    \centering
    \begin{tabular}{cccc}
    \toprule
    \multirow{ 2}{*}{Model} & Baseline & Skip-frame & \multirow{ 2}{*}{Speed-up} \\
    & (ms) & (ms) & \\
    \midrule

    R-FCN  & \multirow{ 2}{*}{1094} & \multirow{ 2}{*}{254} & \multirow{ 2}{*}{4.3$\times$} \\
    ResNet101 & & & \\
       EfficientDet-D0   & 227 & 56 & 4.1$\times$ \\
      \midrule
    SiamFC  & 114 & 52 & $2.8\times$\\
    SiamRPN-2x  & 232 & 65 & 3.6$\times$ \\
    \bottomrule
    \end{tabular}
    \vspace{5pt}
    \caption{Model time and speed-ups achieved by PatchNet on Jetson Nano. Keyframe interval is 5 for video object detection, and 3.6 \& 4.2 for SiamFC \& SiamRPN-2x, following previous experimental settings.
    }
    \label{tab:nano}
\end{table}

\section{Conclusion}
This paper describes PatchNet, a very small neural network model for template matching in short-range video frames, and demonstrates its effectiveness in video object detection and visual object tracking. On both tasks, PatchNet achieves 3-5x speedup with little accuracy drop without retraining the model on the target dataset. As recent advances drive the need for fast video understanding on mobile platforms and compact models like EfficientDet-D0 are developed, previous heavyweight temporal acceleration methods find difficulty achieving speedup. PatchNet provides a viable solution that offers significant and effortless speed-ups on video tasks.

{\small
\bibliographystyle{ieee_fullname}
\bibliography{egbib}
}

\end{document}